\documentclass[11pt]{article}

\usepackage[utf8]{inputenc}
\usepackage[T1]{fontenc}
\usepackage{lmodern}
\usepackage[margin=1in]{geometry}
\usepackage{graphicx}
\usepackage{times}
\usepackage{microtype}
\usepackage{hyperref}
\usepackage{url}
\usepackage{tikz}
\usetikzlibrary{arrows.meta, positioning}

\makeatletter
\renewenvironment{abstract}
  {\noindent\textbf{\abstractname}\par\noindent\ignorespaces}
  {\par}
\makeatother

\title{The LLM Fallacy: Misattribution in AI-Assisted Cognitive Workflows}

\author{
Hyunwoo Kim \quad Harin Yu \quad Hanau Yi \\
ddai Inc. \\
\texttt{hw.kim@ai-dda.com}
}

\date{}

\begin{document}

\twocolumn[
\maketitle

\begin{abstract}
The rapid integration of large language models (LLMs) into everyday workflows has transformed how individuals perform cognitive tasks such as writing, programming, analysis, and multilingual communication. While prior research has focused on model reliability, hallucination, and user trust calibration, less attention has been given to how LLM usage reshapes users’ perceptions of their own capabilities. This paper introduces the LLM fallacy, a cognitive attribution error in which individuals misinterpret LLM-assisted outputs as evidence of their own independent competence, producing a systematic divergence between perceived and actual capability. We argue that the opacity, fluency, and low-friction interaction patterns of LLMs obscure the boundary between human and machine contribution, leading users to infer competence from outputs rather than from the processes that generate them. We situate the LLM fallacy within existing literature on automation bias, cognitive offloading, and human--AI collaboration, while distinguishing it as a form of attributional distortion specific to AI-mediated workflows. We propose a conceptual framework of its underlying mechanisms and a typology of manifestations across computational, linguistic, analytical, and creative domains. Finally, we examine implications for education, hiring, and AI literacy, and outline directions for empirical validation. We also provide a transparent account of human--AI collaborative methodology. This work establishes a foundation for understanding how generative AI systems not only augment cognitive performance but also reshape self-perception and perceived expertise.
\end{abstract}

\vspace{1em}
]

\section{Introduction}

The rapid integration of large language models (LLMs) into everyday workflows has reshaped how individuals perform cognitive tasks, including writing, programming, analysis, and multilingual communication (Achiam et al., 2023). Beyond incremental productivity gains, this shift reflects a structural change in how cognitive labor is organized, with generative systems functioning as embedded components of knowledge work rather than external tools (Brynjolfsson et al., 2025). LLMs do not merely augment isolated tasks but alter the conditions under which problem solving and content generation occur.

Existing research has primarily focused on system-level concerns such as model reliability, hallucination, and user trust calibration (Ji et al., 2023). Parallel work on alignment and human--AI interaction has emphasized improving system behavior through interpretability, controllability, and responsiveness to user intent (Ouyang et al., 2022). While these directions provide important insights into model performance and interaction design, they remain largely system-centered. Comparatively less attention has been given to how sustained interaction with LLMs reshapes users’ perceptions of their own capabilities, particularly in contexts where outputs are co-produced through iterative human--AI exchange.

This paper introduces the concept of the LLM fallacy, defined as a cognitive attribution error in which individuals misinterpret LLM-assisted outputs as evidence of their own independent competence, producing a systematic divergence between perceived and actual capability. The phenomenon can be situated within a broader class of attributional distortions in which individuals misjudge the sources of their performance or knowledge (Kruger \& Dunning, 1999). However, unlike prior accounts that emphasize internal limitations in self-assessment, the LLM fallacy arises from the integration of external generative systems into cognitive workflows, creating hybrid environments in which authorship and agency are not readily separable.

As LLMs mediate an increasing share of cognitive processes, this misattribution produces a systematic divergence between perceived and actual capability. These divergences extend beyond individual misjudgment to affect institutional systems that rely on observable outputs as proxies for competence (Espeland \& Sauder, 2007). When outputs can be produced through human--AI collaboration without corresponding internal expertise, evaluation frameworks risk conflating system-assisted performance with independently grounded skill.

We argue that this phenomenon emerges from the interaction of several properties inherent to LLM systems, including output fluency, interactional immediacy, and the opacity of underlying computational processes (Burrell, 2016). High fluency can function as a metacognitive cue, leading users to infer competence from surface-level coherence rather than from the processes that generate it (Reber et al., 2004). At the same time, the abstraction of intermediate computational steps obscures the division of labor between human and system, limiting users’ ability to accurately attribute contributions.

Taken together, these conditions produce a form of attributional ambiguity that is not incidental but structurally embedded within the interaction. The boundary between human contribution and machine generation is progressively reconstructed through repeated use, leading users to internalize system-assisted outputs as reflections of their own ability.

This paper makes the following contributions. First, it formally defines the LLM fallacy as a cognitive attribution error specific to AI-mediated environments. Second, it differentiates this phenomenon from adjacent constructs, including hallucination, automation bias, and cognitive offloading. Third, it proposes a mechanistic account explaining how LLM interaction produces attributional ambiguity. Fourth, it introduces a multi-domain typology of manifestations across computational, linguistic, analytical, creative, epistemic, and professional contexts. Fifth, it analyzes implications for evaluation systems, including hiring, education, and expertise signaling. Sixth, it provides a structured account of human--AI collaborative methodology to ensure transparency in research practice. Finally, it outlines testable hypotheses and directions for empirical validation.

\section{Background and Related Work}

Research on human interaction with automated systems has long examined how reliance on machine-generated outputs shapes human judgment and behavior (Parasuraman \& Riley, 1997). A central concept in this literature is automation bias, which describes the tendency to over-rely on automated systems, often accepting outputs even when they are incorrect (Lee \& See, 2004). This work shows that reliance is not purely functional but is mediated by perceived system authority and user trust, particularly under conditions of cognitive load.

Closely related is the concept of cognitive offloading, in which individuals externalize mental processes by relying on external systems to perform tasks that would otherwise require internal effort (Risko \& Gilbert, 2016). While offloading reduces cognitive burden, it also reshapes how knowledge is encoded and retained, often diminishing internalization (Hutchins, 1995). As cognitive responsibilities shift outward, the boundary between internal cognition and external support becomes increasingly fluid.

The extended mind framework further develops this perspective by proposing that cognitive processes can be distributed across tools and environments rather than confined to the individual (Clark, 2010). Under this view, technologies do not merely assist cognition but participate in it, forming integrated systems in which reasoning and problem solving are co-constructed. This provides a theoretical basis for understanding LLMs as components within distributed cognitive architectures rather than as external aids.

More recent work in human--AI collaboration examines how users engage with AI systems as partners in task execution (Amershi et al., 2019). These studies emphasize coordination, transparency, and the calibration of user expectations, showing that effective collaboration depends on an accurate understanding of system capabilities and limitations (Kocielnik et al., 2019). However, users often struggle to calibrate trust appropriately, particularly when outputs are fluent or authoritative in presentation.

Research in human-centered AI design further highlights the role of transparency and interpretability in shaping user understanding (Shneiderman, 2020). When system processes are opaque, users are more likely to form incomplete or inaccurate mental models of how outputs are generated, increasing the risk of misinterpretation. Opacity thus functions not only as a technical constraint but as a cognitive condition influencing how system contributions are perceived.

While these frameworks provide important insights into reliance, delegation, and distributed cognition, they do not fully account for the attributional dynamics introduced by LLM-mediated workflows. Existing work focuses primarily on how users evaluate system outputs or integrate them into decision-making, rather than how such outputs are internalized as reflections of personal capability. The question of how externally generated content becomes incorporated into self-assessment remains underexplored.

The LLM fallacy extends this body of work by introducing a distinct form of attributional distortion. Rather than focusing on over-reliance or trust in system correctness, it addresses how LLM-assisted outputs are misattributed as evidence of personal competence. This reframes the analytical focus from interaction and decision-making to self-perception and capability attribution, positioning the phenomenon as a complementary but distinct construct within the broader landscape of human--AI interaction.

\section{Conceptual Definition of the LLM Fallacy}

The LLM fallacy is defined as a cognitive attribution error in which individuals misinterpret LLM-assisted outputs as evidence of their own independent competence. This occurs when system contributions are cognitively absorbed into the user’s self-assessment, producing a misalignment between actual and perceived capability (Kruger \& Dunning, 1999). More broadly, the phenomenon can be understood as a failure of metacognitive monitoring, in which individuals are unable to accurately assess the sources and limits of their own knowledge (Wilson \& Dunn, 2004).

For the LLM fallacy to occur, several conditions must be met. First, the task must involve LLM-mediated output generation, where the system produces content that would otherwise require domain expertise. Second, the interaction must be sufficiently seamless that the distinction between human input and system output is not salient. Third, the output must exhibit a level of fluency or coherence typically associated with skilled human performance (Alter \& Oppenheimer, 2009). Under these conditions, users are more likely to rely on surface-level cues as proxies for competence rather than evaluating the underlying generative process (Reber et al., 2004).

It is important to distinguish the LLM fallacy from related phenomena, particularly hallucination. Hallucination refers to cases in which a model produces incorrect or fabricated information, representing a failure at the level of system output (Ji et al., 2023). The LLM fallacy, by contrast, is independent of output correctness and instead concerns how outputs are cognitively interpreted. It persists regardless of whether generated content is accurate or erroneous, as it operates at the level of attribution rather than epistemic validity.

The LLM fallacy is also distinct from automation bias and cognitive offloading. Automation bias involves over-reliance on system outputs, while cognitive offloading involves delegating mental effort to external systems (Risko \& Gilbert, 2016). Both focus on task execution and decision-making processes. The LLM fallacy instead concerns how outputs are integrated into the user’s self-perception of competence, extending beyond reliance into the domain of capability attribution.

At its core, the phenomenon reflects an attributional misalignment between human and system contributions. In LLM-mediated workflows, the boundary between user input and system-generated output becomes increasingly opaque, making their respective roles difficult to disentangle (Burrell, 2016). This opacity limits the user’s ability to construct accurate mental models of the generative process, increasing reliance on inferred rather than observed causality (Nisbett \& Wilson, 1977). As a result, users may disproportionately attribute outputs to themselves, even when generation is largely system-driven. This misalignment defines the LLM fallacy and provides the foundation for the mechanisms and manifestations examined in subsequent sections.

\section{Mechanism of Emergence}

The LLM fallacy emerges from the interaction of multiple cognitive and system-level factors that jointly produce attributional ambiguity in LLM-mediated workflows. These factors reinforce one another, creating conditions under which users systematically misattribute system-generated outputs as reflections of their own competence (Sloman, 1996). From a dual-process perspective, such misattributions arise when fast, intuitive judgments dominate reflective evaluation, allowing surface-level cues to guide inference.

A primary mechanism is attribution ambiguity between human input and model output. In LLM interactions, users provide prompts that are often partial, underspecified, or iterative, while the system produces structured and coherent outputs. Because results emerge through a continuous interaction loop, the boundary between user contribution and system generation becomes difficult to delineate. This ambiguity increases the likelihood that users incorporate outputs into their sense of authorship, constructing post hoc accounts of their role despite limited introspective access to underlying processes (Nisbett \& Wilson, 1977). Research on agency further shows that authorship is often inferred from outcomes rather than directly accessed, leading to systematic illusions of control and contribution (Aarts et al., 2005). In human--AI contexts, this dynamic is amplified: users may not fully experience ownership of generated content at a cognitive level yet still declare authorship at a reflective or social level, revealing a divergence between experienced and attributed authorship (Draxler et al., 2024). Similar dissociations appear in skilled action, where individuals attribute outcomes to themselves despite incomplete awareness of the processes that produced them (Logan \& Crump, 2010).

A second mechanism is the fluency illusion produced by high-quality natural language generation. LLM outputs are typically grammatically correct, contextually appropriate, and stylistically consistent, closely resembling skilled human performance. This surface-level fluency functions as a heuristic cue, leading users to infer competence from ease of processing rather than from the generative process (Reber et al., 2004). Fluency also biases judgments of credibility and expertise, increasing the likelihood that outputs are perceived as accurate and skillfully produced even in the absence of deeper evaluation (Metzger \& Flanagin, 2013).

Cognitive outsourcing further contributes to the phenomenon. LLMs allow users to externalize complex tasks, including reasoning, composition, and problem solving, with minimal effort. As the system assumes a greater share of cognitive workload, users engage less with the processes required to produce outputs, weakening their ability to assess their own understanding or skill (Kirsh, 2010). Repeated reliance reduces opportunities for self-generated reasoning, reinforcing the gap between perceived and actual competence.

Another critical factor is pipeline opacity, referring to the invisibility of the processes that generate LLM outputs. Unlike traditional tools, where intermediate steps are observable or user-driven, LLMs abstract away retrieval, pattern matching, and synthesis. This opacity prevents users from tracing how outputs are produced and obscures the distinction between system-driven and user-driven contributions (Ananny \& Crawford, 2018). In the absence of transparent intermediate steps, users rely on incomplete mental representations of the system, increasing the likelihood of attribution errors.

Taken together, these mechanisms produce perceived competence inflation. Attribution ambiguity obscures authorship, fluency signals capability, cognitive outsourcing reduces reflective engagement, and pipeline opacity removes visibility into the generative process. Their interaction creates a structurally reinforced environment in which users are consistently inclined to overestimate their own independent competence, giving rise to the LLM fallacy. Formally, this relationship can be summarized as follows: capability divergence ($\Delta C$, defined as the gap between perceived and actual capability) emerges from the interaction of system-level properties, namely opacity, fluency, and immediacy, mediated by attribution ambiguity and cognitive outsourcing. 

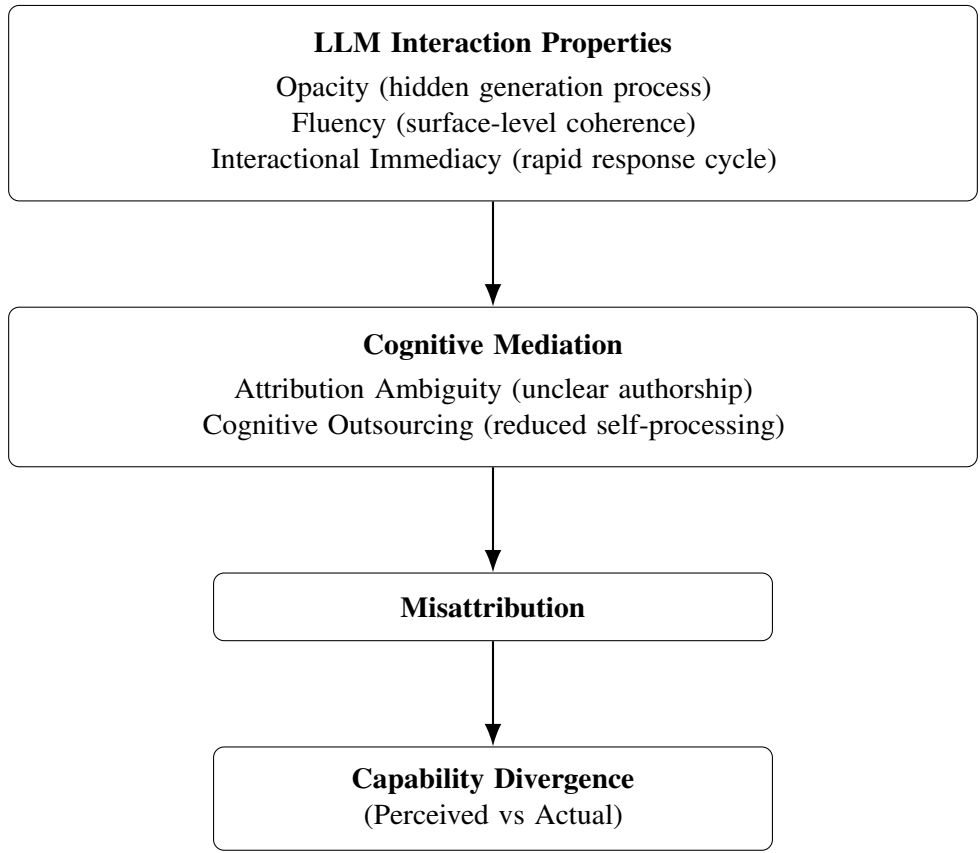
\begin{figure*}[t]
\centering
\begin{tikzpicture}[
    node distance=1.4cm,
    every node/.style={align=center},
    box/.style={
        draw,
        rectangle,
        rounded corners,
        minimum width=0.78\textwidth,
        text width=0.72\textwidth,
        minimum height=1cm,
        inner sep=10pt
    },
    smallbox/.style={
        draw,
        rectangle,
        rounded corners,
        minimum width=0.45\textwidth,
        text width=0.38\textwidth,
        minimum height=0.9cm,
        inner sep=8pt
    },
    arrow/.style={-{Latex[length=3mm]}, thick}
]

\node[box] (interaction) {\textbf{LLM Interaction Properties}\\[0.3em]
Opacity (hidden generation process)\\
Fluency (surface-level coherence)\\
Interactional Immediacy (rapid response cycle)};

\node[box, below=of interaction] (mediation) {\textbf{Cognitive Mediation}\\[0.3em]
Attribution Ambiguity (unclear authorship)\\
Cognitive Outsourcing (reduced self-processing)};

\node[smallbox, below=of mediation] (misattribution) {\textbf{Misattribution}};

\node[smallbox, below=of misattribution] (divergence) {\textbf{Capability Divergence}\\
(Perceived vs Actual)};

\draw[arrow] (interaction) -- (mediation);
\draw[arrow] (mediation) -- (misattribution);
\draw[arrow] (misattribution) -- (divergence);

\end{tikzpicture}
\caption{Mechanism of the LLM fallacy. LLM interaction properties (opacity, fluency, and interactional immediacy) shape cognitive mediation processes, including attribution ambiguity and cognitive outsourcing. These processes produce systematic misattribution of system-generated outputs, resulting in divergence between perceived and actual capability.}
\label{fig:llm_fallacy_mechanism}
\end{figure*}

As illustrated in Figure~\ref{fig:llm_fallacy_mechanism}, these interacting components collectively produce capability divergence through mediated attribution processes.

\section{Typology of LLM Fallacy Manifestations}

The LLM fallacy manifests across multiple domains in which LLMs are used to perform cognitively demanding tasks. While the underlying attributional mechanism remains consistent, its form and visibility vary by task and output type. This section outlines key domains in which the phenomenon is most salient, illustrating how perceived competence inflation emerges across contexts (Dunning, 2011). Across domains, the common structure is a dissociation between externally supported performance and internally grounded understanding.

In the computational domain, the LLM fallacy appears in coding and systems building, where users generate functional scripts or applications with LLM assistance. Users may produce working outputs without understanding underlying architecture, dependencies, or optimization strategies. Execution is thus misinterpreted as evidence of technical competence (Newell \& Simon, 1976), reflecting a distinction between externally scaffolded performance and internally developed understanding.

In the linguistic domain, the phenomenon emerges in multilingual production, where users generate fluent text in languages they do not independently command. Because LLM outputs are often grammatically accurate and contextually appropriate, users may conflate fluency with internalized language ability, overestimating their capacity to comprehend or produce language without assistance (Bender \& Koller, 2020). This reflects a gap between surface form and semantic competence.

In the analytical domain, the LLM fallacy manifests in reasoning and problem-solving tasks. LLMs can generate structured explanations and step-by-step analyses, which users may adopt or reproduce. Exposure to such outputs can create the impression of possessing analytical skill, even when the underlying reasoning process is externally generated (Evans, 2008). Users may internalize the structure of reasoning without engaging in the processes required to produce it independently.

In the creative domain, the phenomenon appears in writing, ideation, and content generation. LLMs assist in producing narratives, arguments, and stylistically refined text, which users may incorporate into their work. These outputs may be misattributed as evidence of personal creativity or authorship despite substantial system contribution (Latour, 1999), reflecting a redistribution of creative agency across human and machine.

In the epistemic domain, the LLM fallacy is observed in knowledge acquisition and understanding. LLMs can summarize complex materials and generate accessible explanations, leading users to equate access to information with conceptual mastery. This aligns with the illusion of explanatory depth, in which individuals overestimate their understanding of complex systems (Rozenblit \& Keil, 2002).

In the domain of professional signaling, the phenomenon manifests in how individuals represent their capabilities in external contexts such as resumes or interviews. Users may report skills based on their ability to produce outputs with LLM assistance rather than independently acquired expertise, resulting in inflated representations of competence that may not transfer to unaided performance (Lamont, 2012).

Together, these domains illustrate the breadth of the LLM fallacy across forms of cognitive work. Despite variation in task type, each domain reflects the same underlying pattern: the conflation of system-assisted output with internally grounded competence.

\section{Empirical Observations and Illustrative Cases}

This section provides observational grounding for the LLM fallacy by identifying recurring patterns across real-world usage contexts. Rather than presenting anecdotal accounts or controlled experimental findings, it focuses on consistent, cross-domain observations that illustrate how the phenomenon manifests in practice (Rosenblat \& Stark, 2016). Such approaches are particularly appropriate for early-stage conceptual work, where stable patterns across contexts can indicate an underlying cognitive structure. These observations are intended as conceptual and cross-contextual patterns rather than as controlled empirical validation.

One prominent pattern appears in coding and system-building workflows. Users can generate functional scripts, applications, or system components through iterative interaction with LLMs, often without understanding the underlying logic, architecture, or debugging processes. While outputs may be operationally effective, users frequently lack the ability to independently reproduce, modify, or extend them. Empirical studies show that LLMs can improve task completion and scaffold code understanding, yet users often rely on generated solutions without internalizing the reasoning behind them (Nam et al., 2024). Moreover, evaluations of code generation systems demonstrate that surface-level correctness does not reliably indicate deeper correctness, as outputs may contain latent errors undetectable without domain expertise (Zhang et al., 2025). This pattern reflects a clear divergence between execution capability and underlying competence, consistent with the definition of the LLM fallacy (Brynjolfsson et al., 2025).

A second pattern emerges in hiring and candidate evaluation contexts. Individuals may present high-quality outputs that reflect LLM-assisted production rather than independently developed skill. When evaluated under conditions requiring unaided performance or deeper understanding, discrepancies become apparent. This divergence is reinforced by evidence that AI assistance can improve observable task performance while increasing reliance on external systems, without corresponding gains in independent capability (Karny et al., 2024). As a result, both self-perception and external evaluation may be shaped by outputs that do not accurately reflect underlying competence, leading to systematic overestimation (Espeland \& Sauder, 2007).

In educational contexts, similar dynamics arise. AI assistance can reduce the need for sustained cognitive engagement, particularly in tasks involving explanation, synthesis, or problem solving. Studies indicate that when AI systems provide solutions or intermediate reasoning, users engage less deeply with the material, limiting opportunities for incidental learning and knowledge internalization (Gajos \& Mamykina, 2022). Although such assistance can improve short-term performance, it weakens the relationship between task completion and conceptual understanding, complicating the interpretation of performance as evidence of learning.

Across domains, a consistent pattern emerges: the ability to produce outputs through LLM interaction is often interpreted as evidence of internalized skill. This aligns with broader findings in cognitive psychology showing that external assistance can induce overconfidence even when underlying understanding remains limited (Fisher \& Oppenheimer, 2021). Such miscalibration reflects a general tendency to infer competence from outcomes rather than from the processes that generate them, a bias that persists even when individuals recognize the role of external aids (Sieck \& Arkes, 2005). In this sense, the LLM fallacy can be understood as a specific instantiation of a broader cognitive bias, amplified by the fluency and immediacy of AI-generated outputs.

More broadly, these observations indicate a shift in how task performance is distributed between human and system. In human--machine teaming, performance emerges from interaction rather than from the isolated capabilities of either component (Damacharla et al., 2018). However, evaluation practices often fail to account for this distribution, treating outputs as attributable to the individual. This misattribution is reinforced by reliance on decision aids, where users defer to system outputs and reduce independent verification and self-assessment (Van Dongen \& Van Maanen, 2013).

Together, these cases demonstrate how the LLM fallacy operates across applied contexts, providing an observational evidence layer that complements the conceptual and mechanistic framework established earlier. Rather than relying on controlled validation, these patterns highlight consistent discrepancies between output quality and underlying competence, bridging abstract definition and real-world manifestation and motivating the need for systematic empirical investigation.

\section{Implications for Evaluation Systems}

The LLM fallacy has implications that extend beyond individual cognition, affecting institutional systems that rely on accurate assessment of human capability. As LLM-assisted workflows become more prevalent, existing evaluation frameworks risk becoming misaligned with the competencies they are intended to measure, particularly when observable outputs are used as primary indicators of skill (Muller, 2018). When outputs can be generated or substantially shaped through AI mediation, the relationship between performance and underlying capability becomes increasingly difficult to interpret, weakening the reliability of outcome-based evaluation as a proxy for competence.

In hiring and candidate assessment, this misalignment introduces a divergence between demonstrated output and independently grounded expertise. Candidates may present work products or articulate solutions that reflect LLM-assisted generation rather than internalized knowledge or skill. Evaluation systems that rely on observable outputs may therefore overestimate competence, especially in contexts where system use is not visible or accounted for (Lamont, 2012). This problem is further compounded by the instability of evaluation itself under AI mediation. Recent work shows that LLM-based evaluation systems can be sensitive to surface-level linguistic features, such as expressions of uncertainty, leading to inconsistent or biased judgments that do not reliably reflect underlying quality (Lee et al., 2025). As a result, both human and automated evaluators may be influenced by features that are orthogonal to actual competence, amplifying the risk of misinterpretation.

In educational contexts, LLM availability alters both learning processes and assessment validity. Students may rely on LLMs to generate explanations, complete assignments, or scaffold problem-solving, reducing the need to engage directly with underlying material. While such tools can enhance access and support learning, they also complicate the interpretation of performance outcomes. Assessment results may no longer reliably reflect conceptual understanding or skill acquisition, creating tension between accessibility and the accurate evaluation of learning progression (Kizilcec, 2016). This challenge is closely related to broader findings that proxy tasks and subjective evaluation measures can produce misleading signals of system effectiveness or user competence, particularly when evaluators rely on superficial indicators rather than underlying processes (Buçinca et al., 2020). In this context, educational assessment faces the risk of conflating assisted performance with genuine understanding.

Similar challenges arise in skill certification and expertise validation systems. Credentials are designed to signal verified competence, yet LLM-assisted workflows enable individuals to meet output-based criteria without possessing corresponding internalized expertise. This weakens the reliability of credentials as indicators of ability, particularly in domains where AI assistance is readily available (Porter, 1995). More broadly, the increasing integration of AI systems into task execution calls into question whether existing evaluation models, which are typically designed for individual performance, remain appropriate for hybrid human--machine settings. Research on human--machine teaming highlights the need for evaluation frameworks that explicitly account for the distribution of contribution between human and system, rather than treating outputs as solely attributable to one or the other (Damacharla et al., 2018).

At a broader level, these shifts reflect a transformation in how knowledge and expertise are produced, interpreted, and validated. As LLMs become embedded within cognitive workflows, the boundary between individual cognition and system-assisted output becomes increasingly diffuse, challenging conventional assumptions about authorship, understanding, and knowledge ownership. This transformation extends beyond specific tasks, affecting how institutions define and evaluate competence in environments where human and machine contributions are inherently intertwined.

These challenges highlight the need for updated AI literacy frameworks that explicitly account for the role of LLMs in shaping cognitive processes and outputs. Such frameworks must move beyond tool usage to address metacognitive awareness, enabling users and evaluators to distinguish between system-assisted performance and independently grounded competence (Liao et al., 2020). This includes developing shared norms around appropriate use, transparency, and interpretation of AI-mediated outputs, as well as clarifying when outputs can and cannot be treated as indicators of skill.

Taken together, these implications indicate that the LLM fallacy operates not only at the level of individual perception but also at the level of institutional evaluation systems. Addressing this misalignment will require rethinking how competence is defined, measured, and communicated in AI-mediated environments, shifting from output-centric models toward process-aware evaluation frameworks that more accurately capture the distribution of human and machine contributions (Power, 1997).

\section{Human--AI Collaborative Methodology and Disclosure}

This study was conducted using a human--AI collaborative workflow in which large language models (LLMs) were used as assistive tools for drafting support, structural refinement, language optimization, and iterative conceptual exploration. All interactions with the LLM were conducted through structured prompts and evaluated by the human author within a controlled interaction framework. In this context, LLMs were treated as systems whose outputs are shaped through interaction design rather than autonomous execution (Bommasani et al., 2021). Outputs were considered intermediate artifacts subject to human interpretation and validation, rather than authoritative contributions.

All conceptual framing, theoretical claims, interpretations, and final decisions were made by the human author, who functioned as the primary investigator and retained full authority over the direction, validation, and integrity of the research. The LLM did not function as an independent author or co-author, nor as a generative authority, but as an assistive system operating under explicit human guidance. This distinction preserves the boundary between language generation and epistemic responsibility, ensuring that authorship remains grounded in human accountability (Huang et al., 2025).

LLM interaction in this study was governed through a structured prompting methodology based on the Natural Language Declarative Prompting (NLD-P) framework (Kim et al., 2026), which was applied to enforce constraints on task structure, output scope, and revision criteria. Unlike conventional prompt engineering approaches that rely on ad hoc instruction crafting or loosely structured interaction patterns (Sahoo et al., 2024), this framework was used here as a control mechanism to ensure that model behavior remained bounded, interpretable, and responsive to explicitly defined requirements. This approach aligns with established principles in human--AI interaction design, where system behavior is shaped through coordinated interaction between user intent and system response (Amershi et al., 2019).

Within this framework, prompts were formulated as modular, human-readable instructions that externalized task definitions and evaluation conditions, enabling consistent application across iterative interactions. This reflects broader efforts to formalize prompting as a deliberate and user-centered practice (Lo, 2023) and supports a shift toward more explicitly coordinated human--AI collaboration (Kraljic \& Lahav, 2024). In this study, these properties were used to maintain clarity regarding the division of roles between human and system, rather than to extend or generalize the framework itself.

All outputs produced through LLM interaction were subject to systematic human review, verification, and selective integration. This included evaluation for conceptual accuracy, logical consistency, and alignment with the intended theoretical framework. Outputs that did not meet these criteria were revised, discarded, or restructured through further interaction. Such iterative validation reflects established practices in AI system auditing and oversight, where human intervention is required to ensure reliability and alignment in system-assisted workflows (Raji et al., 2020). This process also addresses potential discrepancies between system-generated outputs and user-level understanding, consistent with emerging discussions of epistemic alignment in human--LLM interaction (Clark et al., 2025).

This workflow reflects a human-in-the-loop, human-in-control, and human-as-final-author model of collaboration. It aligns with established expectations for transparency in AI-assisted research and explicitly distinguishes between assistive tool usage and authorship, consistent with reporting standards for AI systems (Mitchell et al., 2019). In this context, the structured prompting methodology functions as an operational mechanism for enforcing constraint, maintaining interpretability, and preserving epistemic boundaries, ensuring that LLM-assisted outputs do not obscure the distinction between generated content and human-authored knowledge.

\section{Future Work and Research Directions}

The LLM fallacy, as introduced in this paper, invites systematic empirical investigation. Future work should focus on controlled studies that compare perceived competence under LLM-assisted and unaided conditions, enabling direct observation of divergence between perceived and actual capability (Koriat, 1997). Experimental designs can further manipulate the visibility of system assistance to isolate how attribution shifts under varying levels of transparency.

A central direction involves developing measurement frameworks to quantify this divergence. This includes metrics that compare self-assessed competence with independently evaluated performance, as well as task-specific benchmarks that isolate the contribution of LLM assistance. Such approaches would enable the operationalization of the LLM fallacy as a measurable construct rather than an inferred phenomenon (Wilson \& Dunn, 2004). Critically, these frameworks must capture both subjective perception and objective performance to accurately represent the gap.

Longitudinal studies provide an additional avenue for investigation. As LLM usage becomes embedded in everyday workflows, it is important to examine how repeated interaction shapes users’ self-perception over time. Such work can assess whether prolonged exposure amplifies, stabilizes, or attenuates attributional misalignment (Dillon et al., 2025), and whether users develop more accurate mental models of system contribution or become further entrenched in misattribution.

Domain-specific studies should also be conducted across the typology introduced in Section 5. Differences in task structure, evaluation criteria, and feedback mechanisms may influence both the degree and form of the LLM fallacy. Comparative analyses can identify where the phenomenon is most pronounced and reveal domain-specific moderators such as task complexity, feedback availability, and the visibility of intermediate reasoning (Gigerenzer \& Gaissmaier, 2011).

Finally, future research should examine interventions aimed at reducing attributional misalignment. These may include interface designs that make system contributions more explicit, educational approaches that improve user awareness of LLM capabilities and limitations, and evaluation frameworks that distinguish between system-assisted and independent performance. Such interventions would support the transition of the LLM fallacy from a conceptual construct to an empirically grounded research domain (Doshi-Velez \& Kim, 2017), while contributing to the development of human--AI systems that promote both effective performance and calibrated self-assessment.

\section{Conclusion}

This paper introduced the LLM fallacy as a human--AI interaction phenomenon in which LLM-assisted outputs are misattributed as evidence of independent human competence, producing a systematic divergence between perceived and actual capability (Kruger \& Dunning, 1999). By formally defining this construct, the paper establishes a foundation for understanding how attributional misalignment arises in LLM-mediated workflows and extends existing accounts of self-assessment error to contexts where cognition is distributed across human and machine systems.

The analysis identified the mechanisms underlying this phenomenon, including attribution ambiguity, fluency-driven inference, cognitive outsourcing, and pipeline opacity. Together, these mechanisms create conditions under which users overestimate their capabilities despite substantial reliance on system-generated outputs (Alter \& Oppenheimer, 2009). The typology further demonstrated that this pattern generalizes across computational, linguistic, analytical, creative, epistemic, and professional domains, indicating that the LLM fallacy is not domain-specific but structurally consistent across forms of cognitive work.

Beyond individual cognition, the LLM fallacy has implications for institutional systems that depend on accurate assessment of human capability. Its effects extend to hiring, education, skill certification, and AI literacy frameworks, where observable outputs increasingly fail to reflect underlying ability (Muller, 2018). As the relationship between performance and competence weakens, traditional output-based evaluation models become less reliable indicators of expertise.

Within the broader trajectory of generative AI integration, the LLM fallacy reflects a shift from system-centered concerns toward user-centered cognitive dynamics. While prior work has emphasized model behavior, reliability, and alignment, this paper highlights how LLMs reshape human self-perception and perceived expertise (Bommasani et al., 2021). Understanding AI impact therefore requires not only evaluating system performance but also examining how human cognition adapts under sustained interaction.

This work provides a conceptual framework for the LLM fallacy and identifies directions for future research. Empirical studies are needed to validate the phenomenon, develop measurement methodologies, and design interventions that mitigate attributional misalignment. As LLMs become embedded in cognitive workflows, understanding their impact on self-perception and capability assessment will remain essential. More broadly, addressing the LLM fallacy is necessary to ensure that human--AI collaboration improves not only performance but also the accuracy of self-assessment within hybrid cognitive systems.

\section*{References}

\begingroup
\sloppy

\setlength{\parindent}{0pt}
\setlength{\parskip}{0.5em}
\setlength{\hangindent}{1.5em}
\setlength{\hangafter}{1}

Aarts, H., Custers, R., \& Wegner, D. M. (2005). On the inference of personal authorship: Enhancing experienced agency by priming effect information. \textit{Consciousness and Cognition}, 14(3), 439--458.

Achiam, J., Adler, S., Agarwal, S., Ahmad, L., Akkaya, I., Aleman, F. L., ... \& McGrew, B. (2023). GPT-4 technical report. \textit{arXiv preprint arXiv:2303.08774}.

Alter, A. L., \& Oppenheimer, D. M. (2009). Uniting the tribes of fluency to form a metacognitive nation. \textit{Personality and Social Psychology Review}, 13(3), 219--235.

Amershi, S., Weld, D., Vorvoreanu, M., Fourney, A., Nushi, B., Collisson, P., ... \& Horvitz, E. (2019). Guidelines for human-AI interaction. In \textit{Proceedings of the 2019 CHI Conference on Human Factors in Computing Systems} (pp. 1--13).

Ananny, M., \& Crawford, K. (2018). Seeing without knowing: Limitations of the transparency ideal and its application to algorithmic accountability. \textit{New Media \& Society}, 20(3), 973--989.

Bender, E. M., \& Koller, A. (2020). Climbing towards NLU: On meaning, form, and understanding in the age of data. In \textit{Proceedings of the 58th Annual Meeting of the Association for Computational Linguistics} (pp. 5185--5198).

Bommasani, R., Hudson, D. A., Adeli, E., Altman, R., Arora, S., von Arx, S., ... \& Liang, P. (2021). On the opportunities and risks of foundation models. \textit{arXiv preprint arXiv:2108.07258}.

Brynjolfsson, E., Li, D., \& Raymond, L. R. (2025). Generative AI at work. \textit{The Quarterly Journal of Economics}, 140(2), 889--942. \url{https://doi.org/10.1093/qje/qjae044}.

Buçinca, Z., Lin, P., Gajos, K. Z., \& Glassman, E. L. (2020). Proxy tasks and subjective measures can be misleading in evaluating explainable AI systems. In \textit{Proceedings of the 25th International Conference on Intelligent User Interfaces} (pp. 454--464).

Burrell, J. (2016). How the machine ``thinks'': Understanding opacity in machine learning algorithms. \textit{Big Data \& Society}, 3(1), 2053951715622512. \url{https://doi.org/10.1177/2053951715622512}.

Clark, A. (2010). \textit{Supersizing the Mind: Embodiment, Action, and Cognitive Extension}. Oxford University Press.

Clark, N., Shen, H., Howe, B., \& Mitra, T. (2025). Epistemic alignment: A mediating framework for user--LLM knowledge delivery. \textit{arXiv preprint arXiv:2504.01205}.

Damacharla, P., Javaid, A. Y., Gallimore, J. J., \& Devabhaktuni, V. K. (2018). Common metrics to benchmark human-machine teams (HMT): A review. \textit{IEEE Access}, 6, 38637--38655.

Dillon, E. W., Jaffe, S., Immorlica, N., \& Stanton, C. T. (2025). Shifting work patterns with generative AI (No. w33795). \textit{National Bureau of Economic Research}.

Doshi-Velez, F., \& Kim, B. (2017). Towards a rigorous science of interpretable machine learning. \textit{arXiv preprint arXiv:1702.08608}.

Draxler, F., Werner, A., Lehmann, F., Hoppe, M., Schmidt, A., Buschek, D., \& Welsch, R. (2024). The AI ghostwriter effect. \textit{ACM Transactions on Computer-Human Interaction}, 31(2), 1--40.

Dunning, D. (2011). The Dunning–Kruger effect. In \textit{Advances in Experimental Social Psychology} (Vol. 44, pp. 247--296).

Espeland, W. N., \& Sauder, M. (2007). Rankings and reactivity: How public measures recreate social worlds. \textit{American Journal of Sociology}, 113(1), 1--40. \url{https://doi.org/10.1086/517897}.

Evans, J. S. B. (2008). Dual-processing accounts of reasoning, judgment, and social cognition. \textit{Annual Review of Psychology}, 59, 255--278.

Fisher, M., \& Oppenheimer, D. M. (2021). Harder than you think: How outside assistance leads to overconfidence. \textit{Psychological Science}, 32(4), 598--610. \url{https://doi.org/10.1177/0956797620975779}.

Gajos, K. Z., \& Mamykina, L. (2022). Do people engage cognitively with AI? Impact of AI assistance on incidental learning. In \textit{Proceedings of the 27th International Conference on Intelligent User Interfaces} (pp. 794--806). \url{https://doi.org/10.1145/3490099.3511138}.

Gigerenzer, G., \& Gaissmaier, W. (2011). Heuristic decision making. \textit{Annual Review of Psychology}, 62, 451--482.

Huang, B., Chen, C., \& Shu, K. (2025). Authorship attribution in the era of LLMs. \textit{ACM SIGKDD Explorations Newsletter}, 26(2), 21--43.

Hutchins, E. (1995). \textit{Cognition in the Wild}. MIT Press.

Ji, Z., Lee, N., Frieske, R., Yu, T., Su, D., Xu, Y., ... \& Fung, P. (2023). Survey of hallucination in natural language generation. \textit{ACM Computing Surveys}, 55(12), 1--38.

Karny, S., Mayer, L. W., Ayoub, J., Song, M., Su, H., Tian, D., ... \& Steyvers, M. (2024). Learning with AI assistance. In \textit{Proceedings of the ACM Collective Intelligence Conference}.

Kim, H., Yi, H., Bae, J., \& Kim, Y. (2026). Natural Language Declarative Prompting (NLD-P): A Modular Governance Method for Prompt Design Under Model Drift. \textit{arXiv preprint arXiv:2602.22790}. \url{https://doi.org/10.48550/arXiv.2602.22790}.

Kirsh, D. (2010). Thinking with external representations. \textit{AI \& Society}, 25(4), 441--454.

Kizilcec, R. F. (2016). How much information? Effects of transparency on trust in an algorithmic interface. In \textit{Proceedings of the 2016 CHI Conference on Human Factors in Computing Systems} (pp. 2390--2395). \url{https://doi.org/10.1145/2858036.2858402}.

Kocielnik, R., Amershi, S., \& Bennett, P. N. (2019). Will you accept an imperfect AI? Exploring designs for adjusting end-user expectations of AI systems. In \textit{Proceedings of the 2019 CHI Conference on Human Factors in Computing Systems} (pp. 1--14). \url{https://doi.org/10.1145/3290605.3300641}.

Koriat, A. (1997). Monitoring one's own knowledge during study: A cue-utilization approach to judgments of learning. \textit{Journal of Experimental Psychology: General}, 126(4), 349--370. \url{https://doi.org/10.1037/0096-3445.126.4.349}.

Kraljic, T., \& Lahav, M. (2024). From prompt engineering to collaborating. \textit{Interactions}, 31(3), 30--35.

Kruger, J., \& Dunning, D. (1999). Unskilled and unaware of it: How difficulties in recognizing one's own incompetence lead to inflated self-assessments. \textit{Journal of Personality and Social Psychology}, 77(6), 1121--1134. \url{https://doi.org/10.1037/0022-3514.77.6.1121}.

Lamont, M. (2012). Toward a comparative sociology of evaluation. \textit{Annual Review of Sociology}, 38, 201--221.

Latour, B. (1999). \textit{Pandora’s Hope}. Harvard University Press.

Lee, D., Hwang, Y., Kim, Y., Park, J., \& Jung, K. (2025). Are LLM-judges robust to expressions of uncertainty? Investigating the effect of epistemic markers on LLM-based evaluation. In \textit{Proceedings of the 2025 Conference of the Nations of the Americas Chapter of the Association for Computational Linguistics: Human Language Technologies (Volume 1: Long Papers)} (pp. 8962--8984). \url{https://aclanthology.org/2025.naacl-long.452/}.

Lee, J. D., \& See, K. A. (2004). Trust in automation. \textit{Human Factors}, 46(1), 50--80.

Liao, Q. V., Gruen, D., \& Miller, S. (2020). Questioning the AI: Informing design practices for explainable AI user experiences. In \textit{Proceedings of the 2020 CHI Conference on Human Factors in Computing Systems} (pp. 1--15). \url{https://doi.org/10.1145/3313831.3376590}.

Lo, L. S. (2023). The CLEAR path. \textit{Journal of Academic Librarianship}, 49(4), 102720.

Logan, G. D., \& Crump, M. J. (2010). Cognitive illusions of authorship. \textit{Science}, 330(6004), 683--686.

Metzger, M. J., \& Flanagin, A. J. (2013). Credibility and trust of information in online environments: The use of cognitive heuristics. \textit{Journal of Pragmatics}, 59, 210--220. \url{https://doi.org/10.1016/j.pragma.2013.07.012}.

Mitchell, M., Wu, S., Zaldivar, A., Barnes, P., Vasserman, L., Hutchinson, B., Spitzer, E., Raji, I. D., \& Gebru, T. (2019). Model cards for model reporting. In \textit{Proceedings of the Conference on Fairness, Accountability, and Transparency} (pp. 220--229). \url{https://doi.org/10.1145/3287560.3287596}.

Muller, J. (2018). \textit{The Tyranny of Metrics}. Princeton University Press.

Nam, D., Macvean, A., Hellendoorn, V. J., Vasilescu, B., \& Myers, B. A. (2024). Using an LLM to help with code understanding. In \textit{Proceedings of the IEEE/ACM 46th International Conference on Software Engineering} (pp. 1--13). \url{https://doi.org/10.1145/3597503.3639187}.

Newell, A., \& Simon, H. A. (1976). Computer science as empirical inquiry: Symbols and search. \textit{Communications of the ACM}, 19(3), 113--126. \url{https://doi.org/10.1145/360018.360022}.

Nisbett, R. E., \& Wilson, T. D. (1977). Telling more than we can know: Verbal reports on mental processes. \textit{Psychological Review}, 84(3), 231--259. \url{https://doi.org/10.1037/0033-295X.84.3.231}.

Ouyang, L., Wu, J., Jiang, X., Almeida, D., Wainwright, C. L., Mishkin, P., ... \& Lowe, R. (2022). Training language models to follow instructions with human feedback. \textit{Advances in Neural Information Processing Systems}, 35, 27730--27744. \url{https://doi.org/10.52202/068431-2011}.

Parasuraman, R., \& Riley, V. (1997). Humans and automation: Use, misuse, disuse, abuse. \textit{Human Factors}, 39(2), 230--253. \url{https://doi.org/10.1518/001872097778543886}.

Porter, T. M. (1995). \textit{Trust in Numbers: The Pursuit of Objectivity in Science and Public Life}. Princeton University Press.

Power, M. (1997). \textit{The Audit Society}. Oxford University Press.

Raji, I. D., Smart, A., White, R. N., Mitchell, M., Gebru, T., Hutchinson, B., Smith-Loud, J., Theron, D., \& Barnes, P. (2020). Closing the AI accountability gap: Defining an end-to-end framework for internal algorithmic auditing. In \textit{Proceedings of the 2020 Conference on Fairness, Accountability, and Transparency} (pp. 33--44). \url{https://doi.org/10.1145/3351095.3372873}.

Reber, R., Schwarz, N., \& Winkielman, P. (2004). Processing fluency and aesthetic pleasure: Is beauty in the perceiver's processing experience? \textit{Personality and Social Psychology Review}, 8(4), 364--382. \url{https://doi.org/10.1207/s15327957pspr0804_3}.

Risko, E. F., \& Gilbert, S. J. (2016). Cognitive offloading. \textit{Trends in Cognitive Sciences}, 20(9), 676--688.

Rosenblat, A., \& Stark, L. (2016). Algorithmic labor. \textit{International Journal of Communication}, 10.

Rozenblit, L., \& Keil, F. (2002). The misunderstood limits of folk science: An illusion of explanatory depth. \textit{Cognitive Science}, 26(5), 521--562. \url{https://doi.org/10.1207/s15516709cog2605_1}.

Sahoo, P., Singh, A. K., Saha, S., Jain, V., Mondal, S., \& Chadha, A. (2024). Survey of prompt engineering. \textit{arXiv}.

Shneiderman, B. (2020). Human-centered artificial intelligence: Three fresh ideas. \textit{AIS Transactions on Human-Computer Interaction}, 12(3), 109--124. \url{https://doi.org/10.17705/1thci.00131}.

Sieck, W. R., \& Arkes, H. R. (2005). The recalcitrance of overconfidence and its contribution to decision aid neglect. \textit{Journal of Behavioral Decision Making}, 18(1), 29--53. \url{https://doi.org/10.1002/bdm.486}.

Sloman, S. A. (1996). The empirical case for two systems of reasoning. \textit{Psychological Bulletin}, 119(1), 3--22. \url{https://doi.org/10.1037/0033-2909.119.1.3}.

Van Dongen, K., \& Van Maanen, P. P. (2013). A framework for explaining reliance on decision aids. \textit{International Journal of Human-Computer Studies}, 71(4), 410--424. \url{https://doi.org/10.1016/j.ijhcs.2012.10.018}.

Wilson, T. D., \& Dunn, E. W. (2004). Self-knowledge: Its limits, value, and potential for improvement. \textit{Annual Review of Psychology}, 55, 493--518. \url{https://doi.org/10.1146/annurev.psych.55.090902.141954}.

Zhang, Z., Wang, C., Wang, Y., Shi, E., Ma, Y., Zhong, W., Chen, J., Mao, M., \& Zheng, Z. (2025). LLM hallucinations in practical code generation: Phenomena, mechanism, and mitigation. \textit{Proceedings of the ACM on Software Engineering}, 2(ISSTA), 481--503. \url{https://doi.org/10.1145/3728894}.

\endgroup

\end{document}